\documentclass[lettersize,journal]{IEEEtran}
\usepackage{amsmath,amsfonts}
\usepackage{algorithmic}
\usepackage{algorithm}
\usepackage{array}
\usepackage[caption=false,font=normalsize,labelfont=sf,textfont=sf]{subfig}
\usepackage{xcolor}%
\usepackage{textcomp}
\usepackage{stfloats}
\usepackage{url}
\usepackage{verbatim}
\usepackage{graphicx}
\usepackage{cite}
\usepackage{tabularx}
\usepackage{multirow}
\usepackage[normalem]{ulem}

\begin{document}
\title{\textbf{DatUS$^{2}$}: \textbf{Dat}a-driven \textbf{U}nsupervised \textbf{S}emantic \textbf{S}egmentation with Pre-trained Self-supervised Vision Transformer}
\author{Sonal Kumar, Arijit Sur and Rashmi Dutta Baruah}
\maketitle
\begin{abstract}
Successive proposals of several self-supervised training schemes continue to emerge, taking one step closer to developing a universal foundation model. In this process, the unsupervised downstream tasks are recognized as one of the evaluation methods to validate the quality of visual features learned with a self-supervised training scheme. However, unsupervised dense semantic segmentation has not been explored as a downstream task, which can utilize and evaluate the quality of semantic information introduced in patch-level feature representations during self-supervised training of a vision transformer. Therefore, this paper proposes a novel data-driven approach for unsupervised semantic segmentation (DatUS$^{2}$) as a downstream task. DatUS$^{2}$ generates semantically consistent and dense pseudo-annotated segmentation masks for the unlabeled image dataset without using any visual-prior or synchronized data. We compare these pseudo-annotated segmentation masks with ground truth masks for evaluating recent self-supervised training schemes to learn shared semantic properties at the patch level and discriminative semantic properties at the segment level. Finally, we evaluate existing state-of-the-art self-supervised training schemes with our proposed downstream task, i.e.,  DatUS$^{2}$. Also, the best version of DatUS$^{2}$ outperforms the existing state-of-the-art method for the unsupervised dense semantic segmentation task with 15.02\% MiOU and 21.47\% Pixel accuracy on the SUIM dataset. It also achieves a competitive level of accuracy for a large-scale and complex dataset, i.e., the COCO dataset.

\end{abstract}

\section{Introduction}
\IEEEPARstart{S}{elf}-supervised learning empowers a deep-learning model to capture underlying informative features from a large-size unlabeled dataset. It automates machine learning by eliminating pre-training human effort for manual annotation and preprocessing of datasets. Also, it encompasses a wide range of deep learning applications that utilize text, speech, image, or video datasets \cite{stefanov2019self, jing2020self,zhang2023graph,goyal2022vision}. Consequently, many state-of-the-art methods exist for self-supervised training of deep learning models like CNN, GNN, and transformers \cite{chen2020simple,caron2020unsupervised,he2020momentum,fang2022self,vaswani2017attention,devlin2018bert,caron2021emerging,zhou2021ibot,zhou2022mugs,an2023unicom,bar2022detreg}. In the absence of human annotation, the goal of a general self-supervised training scheme is to learn a meaningful kernel to facilitate a wide variety of downstream computer vision tasks like image classification, semantic segmentation, depth estimation, etc. A considerable performance of various downstream tasks verifies the generality of a self-supervised training scheme. Hence, the prevailing approach to validate a self-supervised training scheme is to evaluate the performance of such downstream tasks with the pre-trained model \cite{chen2020simple,caron2020unsupervised,he2020momentum,vaswani2017attention,devlin2018bert,caron2021emerging,zhou2021ibot,zhou2022mugs,an2023unicom}. A downstream task can be a supervised or unsupervised computer vision application. In contrast with the supervised downstream task, an unsupervised task directly uses the feature representation extracted from the self-supervised deep learning model. For instance, supervised downstream tasks like image classification utilize the pre-trained model for further fine-tuning or linear evaluation with the small annotated dataset. By contrast, in unsupervised tasks such as image clustering (unsupervised classification), the image-level feature representations extracted from the pre-trained model are utilized directly for subsequent processing (like k-means clustering)\cite{zheltonozhskii2020self}. Hence, the unsupervised downstream tasks are the best evaluators for the quality of visual features learned during self-supervised training. 

The performance of different unsupervised downstream tasks with a pre-trained model provides insight into the corresponding task-specific properties introduced in visual feature representations during self-supervised pre-training. For example, the performance of image clustering as a downstream task evaluates the self-supervised training to learn image-level or global feature representations. Similarly, an unsupervised dense semantic segmentation task evaluates it to learn pixel-level or local feature representations. However, introducing image and pixel-level discriminative properties during self-supervised training with the CNN model requires separate training schemes \cite{wang2022fully,ji2019invariant}. The introduction of self-supervised training in vision transformers unlocked the true potential of machine learning to develop a general self-supervised training scheme. In contrast with the hierarchical feature extraction of convolution neural network (CNN) architecture, it utilizes the self-attention mechanism to capture the global dependence, which allows it to learn image and patch-level feature representations simultaneously \cite{dosovitskiy2020image}. Unlike the CNN model, image and patch-level feature representations obtained from the same self-supervised training of the vision transformer can be used to perform two different unsupervised downstream tasks, i.e., image clustering and unsupervised dense semantic segmentation, respectively. 

Semantic properties are crucial for describing and understanding visual data more meaningfully. It includes spatial relationships and contextual information within a scene, along with the object categories, attributes, and scene types. The dense semantic segmentation task assigns a label to each pixel of an image so that pixels with a similar label share some common semantic properties throughout the whole dataset \cite{ke2022unsupervised,ji2022semantic}. Although an unsupervised counterpart of image classification, i.e., image clustering using image-level feature representations (CLS tokens), exists to evaluate the self-supervised training scheme in a fully unsupervised setting, an unsupervised dense semantic segmentation task directly using the extracted patch-level feature representations (patch embeddings) from a self-supervised vision transformer is not explored as a downstream task. A few research contributions have been made to explore the potential of patch-level feature representation extracted from self-supervised vision transformers for computer vision tasks like object localization \cite{simeoni2021localizing, simeoni2023unsupervised,wang2022self,lim4251338k} and object mask prediction \cite{van2022discovering} for object-centric datasets. Such contributions suggest that the self-supervised vision transformer learns patch embedding containing high-level semantic and spatial information—these patch embeddings from the same visual groups in an image share high-level semantic relation. In particular, with an appropriate strategy, it is possible to decompose a scene-centric image into multiple visual groups or segments by utilizing the semantic information shared by these patch embeddings. Subsequently, these image segments can be further processed to distill semantically consistent dense pseudo-annotated segmentation masks of images from a scene-centric dataset \cite{vobecky2022drive}. With this idea, we propose a novel data-driven strategy for a self-supervised vision transformer that can distill pseudo-annotated segmentation masks of input images without using any visual-prior, manual annotation, or synchronized data like LiDAR images \cite{vobecky2022drive}. The proposed task can evaluate recent self-supervised training schemes in a fully unsupervised setting to learn shared semantic properties at the patch level and discriminative properties at the segment level.

According to our knowledge, this is the first attempt to propose a dense semantic segmentation as an unsupervised downstream task for self-supervised training schemes of a vision transformer. Our approach is data-driven because it solely works with the feature representation extracted from the pre-trained vision transformer and does not require any further training with the image dataset. In contrast with the existing unsupervised semantic segmentation approaches, the proposed method generates a pseudo-annotated segmentation mask for the entire scene rather than a few centered objects without using any visual prior, human annotation, or synchronized data. The major contributions of this paper are as follows:
\begin{enumerate}
    \item We propose a novel data-driven approach for an unsupervised semantic segmentation, DatUS$^{2}$, to distill dense pseudo-annotated segmentation masks of an image dataset. The key strategy is to utilize the patch embeddings extracted from a pre-trained self-supervised vision transformer to decompose a scene into multiple segments, which is crucial for whole-scene segmentation.
    \item We show that the proposed unsupervised dense semantic segmentation method can be used as a downstream task to evaluate the self-supervised training schemes for the vision transformer in a fully unsupervised setting.
    \item The proposed method DatUS$^{2}$ (ViT-B/8) outperforms the current state-of-the-art, i.e., STEGO (ViT-S/8) on the SUIM dataset with 15.02 \% MiOU and 21.47 \% Pixel accuracy. The performance further improves up to 37.40 \% MiOU and 31.44 \% with the overclustering. It also achieves a competitive level of accuracy on the large-scale dataset, i.e., the COCO dataset.
\end{enumerate}
Alternatively, the proposed method DatUS$^{2}$ along with the additional de-noising step can be seen as a highly informative visual priors mining scheme. Unlike existing low or mid-level visual prior, it consists of class-level information and captures all entities of an input image. The subsequent sections of the paper are structured as follows. Section \ref{sec2} presents a detailed review of the related works. Section \ref{sec3} elucidates the details of the proposed method. Section \ref{sec4} discusses the experiments' datasets, setup, and findings. Section \ref{sec5} presents a range of concerns and potential avenues for future exploration. Finally, section \ref{sec6} concludes the overall research work. 
\section{Related Work} \label{sec2}
\subsection{Self-supervised Training}
Self-supervised training originated in the natural language processing domain \cite{vaswani2017attention} and subsequently found its way into the field of computer vision. 
As a result, many self-supervised training schemes exist with the capability of training both CNN and vision transformer-based deep learning models \cite{chen2020simple,caron2020unsupervised,he2020momentum,devlin2018bert,caron2021emerging,zhou2021ibot,zhou2022mugs,an2023unicom}. 
The self-supervised training scheme can be task-specific or generalized. A generalized self-supervised training scheme provides a strong baseline for downstream tasks. This paper focuses on the generalized self-supervised schemes for training a vision transformer and establishes a novel unsupervised downstream task for dense semantic segmentation to evaluate them. 

The key ingredient of a self-supervised training scheme is the supervisory signal generated from the unlabeled dataset. The continuous dedication of the research community has led to the development of numerous self-supervised training approaches for the vision transformer, each incorporating distinct types of supervisory signals. The authors of \cite{caron2021emerging} propose the DINO framework to study the compatibility of the contrastive approach of self-supervised training in a vision transformer \cite{dosovitskiy2020image}. The observation suggests that self-supervised training introduces a high level of semantic information into the self-attention map obtained from the vision transformer's last block. In the next version of the DINOv2 framework  \cite{oquab2023dinov2}, the authors scale up the pre-training in terms of model and data size. The objective is to enhance the training process to make visual features effective across various data distributions and tasks without fine-tuning. The success of the Mask Language Modeling (MLM) pretext task for the language transformer  \cite{devlin2018bert} motivates the authors of paper \cite{zhou2021ibot} to propose a novel Mask Image Modeling (MIM) pretext task for the vision transformer.
Similarly, a novel self-supervised training scheme introduced in the Mugs framework improves the learning of the vision transformer \cite{zhou2022mugs}. It utilizes the multi-granular feature of input images as the supervisory signal for self-supervised training. The authors of the paper \cite{an2023unicom} identify the limitation of self-supervised training with images from a limited number of classes. Using the CLIP framework \cite{radford2021learning}, they introduce a post-training strategy to partition a large-scale dataset into one million pseudo-classes. Next, they utilize the pseudo-class information with a margin-based softmax function to train the vision transformer.
\subsection{Unsupervised Semantic-segmentation} In the absence of supervision, semantic segmentation can be seen as a clustering operation at the pixel level. That is why it is crucial to learn quality pixel embeddings, which are discriminative enough to form meaningful clusters. Various self-supervised training schemes for deep learning models have been proposed to learn good-quality pixel embeddings. 

A group of work focuses on evolving the end-to-end training approach \cite{ji2019invariant,cho2021picie,hamilton2022unsupervised,harb2022infoseg,hwang2019segsort} in the direction of unsupervised semantic segmentation. The objective is to learn pixel embeddings and pixel labeling simultaneously. In early work, the authors proposed a self-supervised framework, IIC \cite{ji2019invariant}, which learns meaningful pixel embedding by maximizing the mutual information between the cluster assignment of each pair of neighboring patches from input images. The SegSort framework \cite{hwang2019segsort} iteratively performs pixel sorting followed by segment sorting to learn pixel embedding along with the cluster assignment. It first discovers visual segments among pixel embeddings by applying spherical k-means clustering, dubbed pixel sorting, and then proposes an objective function to separate and group the segments into semantic clusters, dubbed segment sorting. The PiCIE framework \cite{cho2021picie} learns meaningful pixel embedding by incorporating geometric and photometric consistency between the pixels of the different views of an image with a contrastive loss function of a non-parametric nature. The subsequent research paper \cite{wang2022fully} induces global-level information in pixel embedding by alternatively performing image-level and pixel-level clustering. The pixel-level clustering leverages the PiCIE \cite{cho2021picie} framework. The STEGO framework \cite{hamilton2022unsupervised} proposes a feature correspondence-based contrastive loss function to train a segmentation head over the frozen CNN backbone of an unsupervised feature learning framework. Unlike these, the InfoSeg framework leverages CNN to extract local features and multiple global features for input images. Next, it introduces an objective function to maximize the mutual information of local and global features.

Despite the significant contribution to the end-to-end training approach, the model performance is sensitive to the random initialization of the model parameters. To overcome this issue, another set of research, i.e., the bottom-up approach, developed parallel to the end-to-training approach. The objective is to utilize a predetermined visual prior for decomposing the scene in perceptual groups and a loss function to leverage the semantic information of such groups to learn the meaningful pixel embedding. The InMARS framework \cite{mirsadeghi2021unsupervised} first decomposes an input image into meaningful regions, i.e., superpixels, and utilizes the mutual information maximization function and adversarial training to learn superpixel's cluster assignments. Similarly, the MaskContrast framework \cite{van2021unsupervised} utilizes an unsupervised saliency estimator \cite{nguyen2019deepusps} to obtain a mid-level visual prior, i.e., object saliency mask and uses a novel contrastive objective function to discriminate pixels from different object saliency masks from multiple images. Unlike this, a few research works \cite{vobecky2022drive,van2022discovering} generate segment-mask proposals of input images using prior knowledge to train the segmentation model from scratch. The Drive\&Segment framework \cite{vobecky2022drive} uses cross-model information of the image and corresponding LiDAR point cloud to generate corresponding segment-mask proposal. Later, train a segmentation model with images and segmentation-mask proposal pair. Unlike that, the MaskDistill framework \cite{van2022discovering} utilizes a vision transformer-based unsupervised object localization framework, i.e., LOST \cite{simeoni2021localizing} to generate foreground object-mask proposal. Later, train a  region proposal network, i.e., \cite{he2017mask} from scratch with image and object-mask proposal pair. 

It is observed in the above discussion that the common goal of most of the best-published work is to establish a foundation framework for self-supervised training of the vision transformers. 
Existing research ensures this by evaluating the proposed self-supervised training scheme with a KNN classifier and supervised or unsupervised downstream computer vision tasks. We observe that the existing downstream tasks lack an unsupervised semantic segmentation performed directly using the extracted features, i.e., patch embeddings from the pre-trained vision transformer. The above literature also suggests that there is scope to establish a bottom-up approach to do the same. 
Based on our observation, we propose a novel unsupervised semantic segmentation technique as an unsupervised downstream task for a self-supervised vision transformer. The proposed method successfully segments scenes into multiple visual groups/segments using the patch embeddings extracted from the pre-trained vision transformer to generate pseudo-annotated segmentation masks for an image dataset. 

\textit{How is our work different from the existing unsupervised semantic segmentation methods?} The existing end-to-end unsupervised semantic segmentation methods perform dedicated task-specific training, which requires a large-scale dataset. Unlike this, we propose a novel bottom-up approach, which generates pseudo-annotated segmentation masks for large as well as small-size datasets without task-specific training. In contrast with the existing bottom-up approaches, our proposed method works upon a vision transformer as a novel unsupervised downstream task of self-supervised training schemes. Unlike the existing approaches, it is a scene-centric or dense segmentation method. Also, mining pseudo-annotated segmentation masks for a given unlabeled image dataset does not require further training and visual priors or synchronized data. 
\section{Proposed Method} \label{sec3}
\begin{figure*}[ht]
\centering
\includegraphics[width=7in]{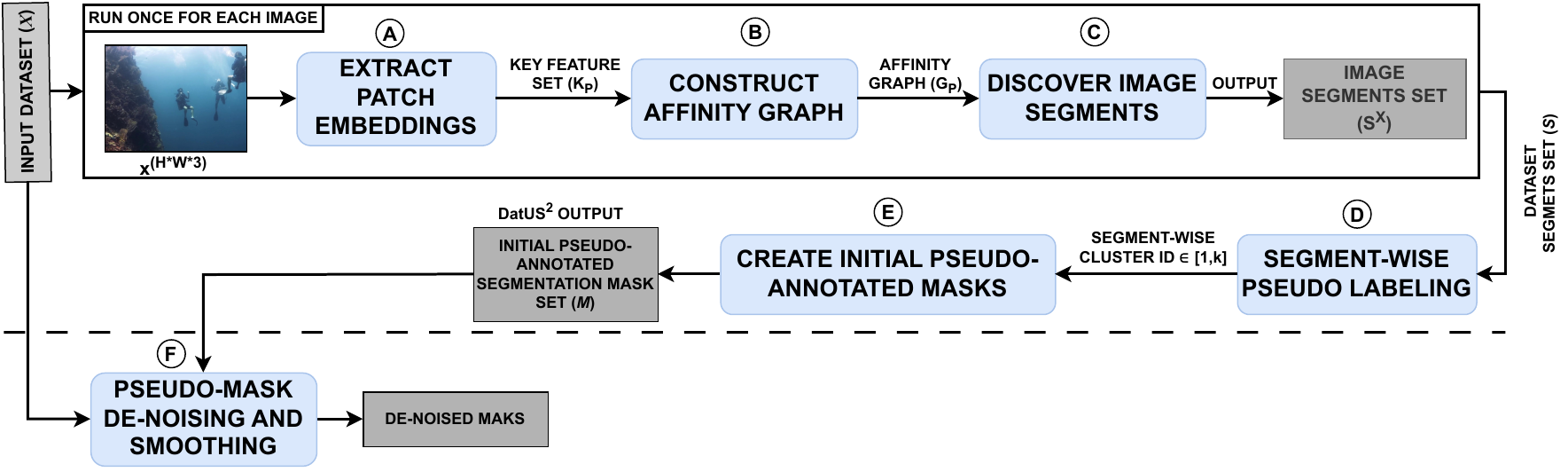}
	\caption{An overview of the proposed method, i.e., data-driven unsupervised semantic segmentation with pre-trained self-supervised vision transformer (DatUS$^{2}$). The first three steps of the proposed method, i.e., \textit{Extract Patch Embeddings}, \textit{Construct Affinity Graph}, and \textit{Discover Image Segments}, operate on a single image at a time. After applying these three steps to each image in the dataset, the method proceeds with the remaining steps, i.e., \textit{Segment-wise Pseudo Labeling}, \textit{Create Initial Pseudo-annotated Masks}, and \textit{Pseudo-mask De-noising and Smoothing}.}
	\label{fig:0}
\end{figure*}

The proposed method, i.e., DatUS$^{2}$, harnesses the potential of vision transformer, self-supervised learning, unsupervised learning, and image processing to mine semantically consistent pseudo-annotated segmentation masks of a scene. The major components of the proposed method are: \textit{Self-supervised Vision Transformer (ViT)}: To extract the patch embeddings for an input image. \textit{Unsupervised Graph Clustering Algorithm}: To discover multiple independent image segments with patch embeddings. \textit{Self-supervised Feature Extractor (CNN/ViT)}: To extract feature representations of discovered segments of input images. \textit{K-means Clustering Algorithm}: To label the segments of input images using segment-wise feature representations. 

Figure \ref{fig:0} summarises the workflow of our proposed method, i.e., DatUS$^{2}$. The major steps of DatUS$^{2}$ are \textit{Extract Patch Embeddings}, \textit{Construct Affinity Graph}, \textit{Discover Image Segments}, \textit{Segment-wise Pseudo Labeling}, and \textit{Create Initial Pseudo-annotated Masks}. Additionally, we introduce \textit{Pseudo-mask De-noising and Smoothing} step for further smoothing of DatUS$^{2}$ output. The first three steps of DatUS$^{2}$ operate on a single image at a time. After applying these three steps to each image in the dataset, the method proceeds with the remaining steps.

\begin{figure}[ht]
\centering
\includegraphics[width=3.45in]{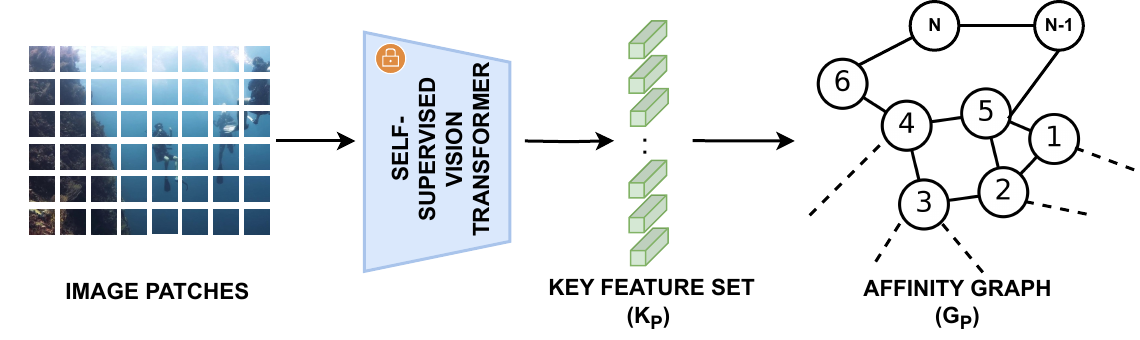}
\caption{Constructing affinity graph using key feature set $K_{P}$ extracted from pre-trained self-supervised vision transformer.}\label{fig:2}
\end{figure}
\subsection{Extract Patch Embeddings}\label{subsecM1} Our approach is based on a vision transformer that is pre-trained in a self-supervised manner. \cite{vaswani2017attention,devlin2018bert,caron2021emerging,zhou2021ibot,zhou2022mugs,an2023unicom}. The vision transformer processes an input image as a sequence of flattened 2D non-overlapping patches called patch tokens, i.e., $\{p_{1},p_{2},p_{3},...,p_{N}\} \epsilon P $. We reshape an input image $X^{(H*W*3)}$ from dataset $\mathcal{X}$ to a dimension of (T, T), i.e., $X^{(T*T*3)}$ and input to the pre-trained vision transformer. The vision transformer decomposes it into N number of patches of dimension (t,t), i.e., $x_{i}^{(t*t*3)}$. Further, it appends a CLS token with the patch tokens, i.e., a total of $N+1$ number of tokens are input to the encoder, where $N = T^{2}/t^{2}$. Each head of the vision transformer's multi-head self-attention (MSA) block computes a feature vector for each non-overlapping patch. We concatenate the feature vectors of multiple heads from the final MSA block of the vision transformer along their embedding dimension and project them into three spaces, i.e., key ($K_{P}$), query ($Q_{P}$), and Value($V_{P}$). 
\subsection{Construct Affinity Graph}\label{subsecM2} We adopt the idea of \cite{simeoni2021localizing} to construct an affinity graph based on the similarity between the image patches as shown in Figure \ref{fig:2}. Unlike them, We use the N number of key feature vectors concatenated over multiple heads, i.e., $K_{P}$, to construct an affinity graph $G_{P}$. The affinity matrix $A_{P} \in \mathcal{R}^{N \times N}$ computes the pairwise similarity between Patch tokens, i.e., $A_{P} = K_{P} \cdot K_{P}^{T}$. The entry $a_{mn}$ of $A_{P}$ matrix gives the dot product of key feature vectors $k_{p_{m}}$ and $k_{p_{n}}$ of patch tokens ${p_{m}}$ and ${p_{n}}$, respectively. Further, $m^{th}$ and $n^{th}$ nodes of the affinity graph $G_{P}$ share a common edge if the $a_{mn}$ entry of $A_{P}$ matrix is positive. In mathematical terms, if the value of an entry in the affinity matrix $A_{P}$ is positive, then the corresponding entry in the adjacency matrix $Adj_{P}$ of an affinity graph $G_{P}$ is one, else zero as shown in equation \ref{e1}.
\begin{align} \label{e1}
Adj_{P} = 
	\begin{cases} 
      adj_{mn} = 0, & a_{mn}\leq 0 \\
      adj_{mn} = 1, & a_{mn} > 0
   \end{cases}
\end{align}

\subsection{Discover Image Segments}\label{subsecM3} Now, given an affinity graph $G_{P}$, we seek to partition the affinity graph into meaningful groups in an unsupervised manner. Experimentally, we discover that the graph partitions of $G_{P}$, obtained from a popular unsupervised graph partition algorithm, i.e., the Louvain algorithm \cite{blondel2008fast}, perfectly align with the object-wise pixel groups of the input image. The Louvain algorithm is a greedy algorithm for community detection in large networks. The Louvain performs hierarchical clustering to maximize the graph modularity $M_{P}$ and gives the optimal number of network partitions. The equation \ref{e2} defines the graph modularity $M_{P}$. 
\begin{align} \label{e2}
M_{P} = 
{
	\frac{1}{2m} \sum_{\alpha\beta} \Bigg[EW_{\alpha\beta} - \frac{TW_{\alpha} TW_{\beta}}{2m}\Bigg] \delta(C_{\alpha},C_{\beta})
}
\end{align}
where $\alpha$ and $\beta$ are two nodes of a network/graph belonging to community/cluster $C_{\alpha}$ and $C_{\beta}$, respectively, $ EW_{\alpha\beta}$ notifies the weight of the edge between $\alpha$ and $\beta$, $TW_{\alpha}$ represents the total weight of edges incoming to node $\alpha$, $m = \sum_{\alpha,\beta}EW_{\alpha\beta}$, and $\delta(C_{\alpha}, C_{\beta})$ is equal to one if $\alpha = \beta$, else zero.
\begin{figure*}[ht]
\centering
\includegraphics[width=5in]{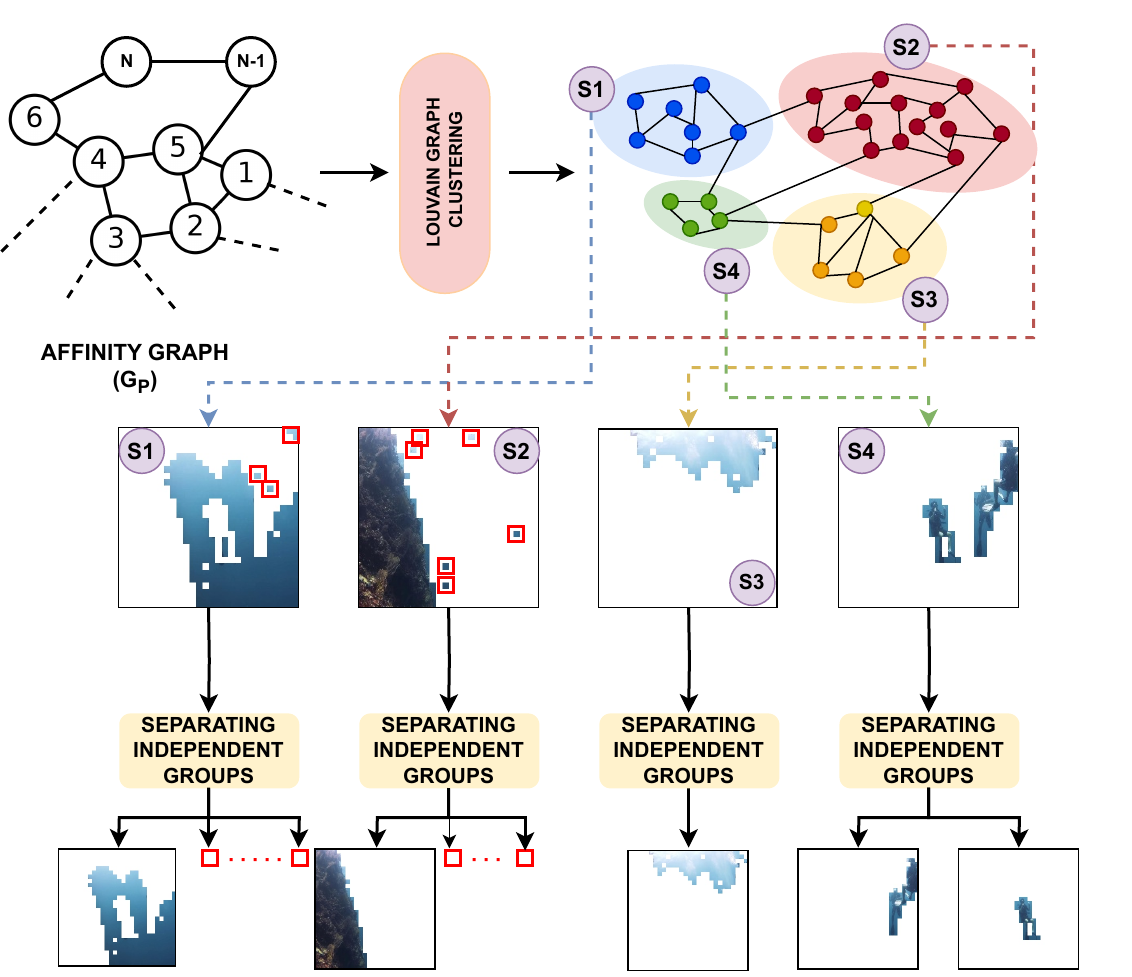}
	\caption{The process of discovering segments from an affinity graph with unsupervised graph clustering, i.e., Louvain Clustering algorithm, followed by the further decomposition step. The red box highlights the noisy sub-segment of resulting segments.}
	\label{fig:3}
\end{figure*}
As shown in Figure \ref{fig:3}, applying the Louvain algorithm to the affinity graph $G_{P}$ results in the $\mathbf{I}$ number of non-overlapping graph partitions (group of nodes). Combining the corresponding patches from each partition decomposes the input image $\mathbf{X}$ into $\mathbf{I}$ number of non-overlapping segments. We observe that some resulting partitions consist of multiple disconnected segments. Hence, we further process the resulting partition of affinity graph $G_{P}$ to decompose them into $\mathbf{J}$ disconnected segments, where $\mathbf{J} \geq \mathbf{I}$. Finally, we extract the $\mathbf{J}$ non-overlapping segments of the image $\mathbf{X}$, i.e., $\mathbf{S^{X}} = \{\mathbf{s^{X}_{j}}\}_{j=1}^{J}$ from $\mathbf{J}$ number of graph partitions. Each segment $\mathbf{s^{X}_{i}}$ is a group of image patches covering some entity in image $\mathbf{X}$. 
\begin{algorithm}[h!]\label{Al1}
\caption{ DatUS$^{2}$ algorithm.}\label{alg:alg1}
\begin{algorithmic}[1]
\STATE \textbf{Input}: Image dataset $\mathcal{X}$, Image dataset size D, Pre-trained Vision Transformer $\mathcal{VT}$, Unsupervised Graph Partition Algorithm $\mathcal{LUV}$, Patch Count Threshold $\mathcal{T}$, Number of Clusters K, Segmentation Model $\mathcal{SM}$.
\STATE $\mathcal{S}$ = \{\}; //Set of image segments
\STATE $\mathcal{M}$ = \{\}; //Set of initial pseudo-annotated segmentation masks
\STATE \textbf{for} d=0 to D do

\STATE \hspace{0.5cm}Resize $X^{(H*W*3)}_{d}$ to $X^{(T*T*3)}_{d}$. 
\STATE \hspace{0.5cm}Feed $X^{(T*T*3)}_{d}$ to pre-trained $\mathcal{VT}$ and extract N+1 patch embeddings from the last MSA block.
\STATE \hspace{0.5cm}Project patch embeddings into three spaces, i.e., key ($K_{P}$), query ($Q_{P}$), and Value($V_{P}$).

\STATE \hspace{0.5cm}Compute Affinity matrix $A_{P} \in \mathcal{R}^{N \times N}$ as a dot product of $K_{P}$ and $K_{P}^{T}$. 
\STATE \hspace{0.5cm}Compute Adjacency matrix $Adj_{P}$ from $A_{P}$ using equation \ref{e1}.

\STATE \hspace{0.5cm}Obtain $\mathbf{I}$ number of non-overlapping graph partitions of $G_{P}$ by feeding $Adj_{P}$ to $\mathcal{LUV}$.
\STATE \hspace{0.5cm}Extract $\mathbf{J}$ disconnected segments of image $\mathbf{X}_{d}$, i.e., $\mathbf{S^{X}} = \{\mathbf{s^{X}_{j}}\}_{j=1}^{J}$ from $\mathbf{I}$ number of graph partitions, where $\mathbf{J} \geq \mathbf{I}$.
\STATE \hspace{0.5cm}Include $\mathbf{S^{X}}$ to $\mathcal{S}$, i.e., $\mathbf{S^{X} \cup \mathcal{S}}$.
\STATE \textbf{end for}

\STATE Generated crop dataset, i.e., $\mathcal{C}$, by cropping the tight rectangle around each image segment (having at least $\mathcal{T}$ number of patches) of each image.
\STATE Feed crops dataset $\mathcal{C}$ from the pre-trained $\mathcal{VT}$ and extract fixed-size feature representation set (CLS tokens), i.e., $\mathcal{F}$, 
\STATE Train a k-means clustering algorithm with the feature representation set $\mathcal{F}$ and obtain the cluster id in [1, K] for each corresponding image segment.

\STATE \textbf{for} i=0 to D do
\STATE \hspace{0.5cm}Produce the pseudo-annotated segmented mask $\mathbf{M_{X_{i}}}$ of image $\mathbf{X_{i}}$ by assigning the cluster id $\mathbf{k}$ of the segment $\mathbf{s^{X}_{j}}$ to the corresponding image pixel.
\STATE \hspace{0.5cm}Include $\mathbf{M_{X_{i}}}$ to $\mathcal{M}$, i.e., $\mathbf{M_{X_{i}}} \cup \mathcal{M}$.
\STATE \textbf{end for}

\STATE Train $\mathcal{SM}$ from scratch with image dataset $\mathcal{X}$ and corresponding pseudo segmentation masks $\mathcal{M}$.
\STATE \textbf{Output} $\mathcal{M}$, $\mathcal{SM}$
\end{algorithmic}
\label{alg1}
\end{algorithm}
\subsection{Segment-wise Pseudo Labeling}\label{subsecM4} 
\begin{figure}[ht]
\centering
\includegraphics[scale=0.44]{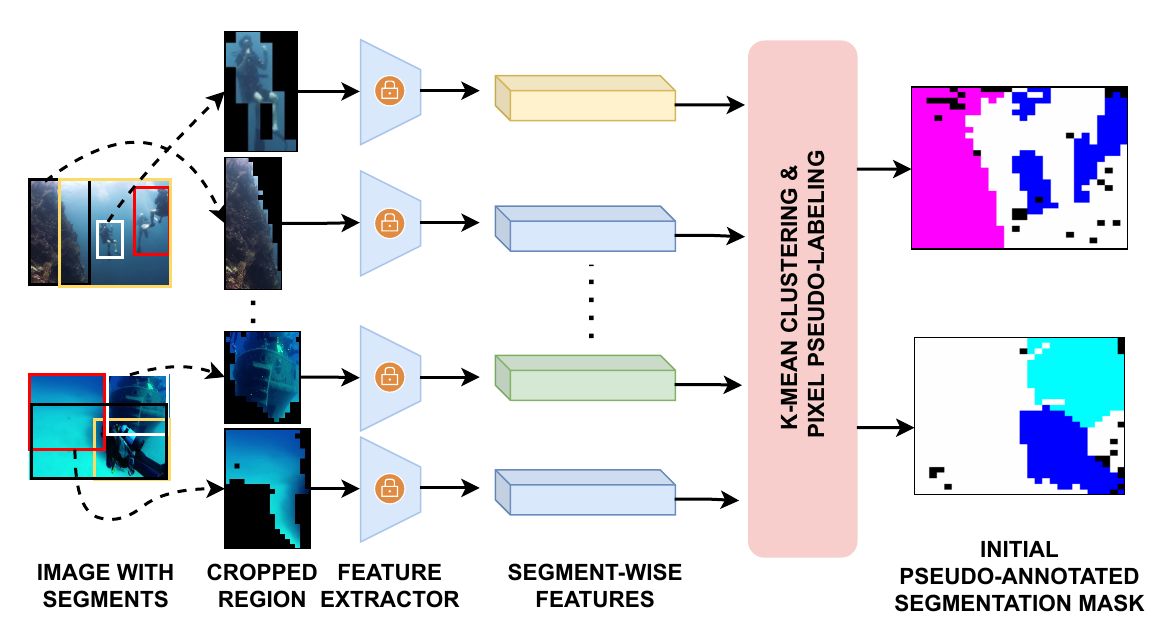}
	\caption{The process of the segment-wise pseudo-labeling to generate \textit{initial pseudo-annotated segmentation masks}.}
	\label{fig:4}
\end{figure}

Now, we have the image dataset $\mathcal{X}$ and its corresponding segments set $\mathcal{S} = \{\mathbf{S^{X_{i}}}\}_{i=1}^{D}$, where D is the size of the image dataset. Our next goal is to extract a set of pseudo-annotated segmentation mask $\mathcal{M} = \{\mathbf{M^{X_{i}}}\}_{i=1}^{D}$ for image dataset $\mathcal{X}$. To do so, we pseudo-labels all extracted segments from the image dataset in an unsupervised manner. We utilize two-stage self-supervised image classification to produce the pseudo labels. Figure \ref{fig:4} explains the process of segment-wise pseudo-labeling to create an initial pseudo-annotated segmentation mask. 
In this work, a segment is a group of image patches covering some entity in an image. We discover the segments in the previous step by applying the Louvain algorithm on the affinity matrix and further decomposing the resulting partition into individual segments. We found the resulting set of segments of an image consists of some noise segments covering only a small region of an entity and made of a very small number of patches. We ignore such segments for the self-supervised pseudo-labeling part to omit their influence. First, we separate all noise segments made of patches less than some pre-determined threshold $\mathcal{T}$ and perform two-stage self-supervised image classification with the rest of the segments.

Following the work of \cite{vobecky2022drive}, we crop the tight rectangle around each segment $\mathbf{s^{X}_{i}} \in \mathbf{S^{X}}$ of an image $\mathbf{X} \in \mathcal{X}$ and mask out the pixels that belong to the other segment of the image. We perform this action on every segment (having at least $\mathcal{T}$ number of patches) within every image of the training dataset to generate a crop dataset, i.e., $\mathcal{C}$. In the first stage, we extract the fixed-size feature representation of all crops by feeding it to the CNN/Transformer-based backbone, which is trained in a self-supervised manner. In the case of the CNN-based backbone \cite{he2020momentum}, we extract feature representation from the last layer, and CLS token in the case of the transformer-based backbone \cite{vaswani2017attention,devlin2018bert,caron2021emerging,zhou2021ibot,zhou2022mugs,an2023unicom}. In the following stage, we train a k-means clustering algorithm with the feature representation set $\mathcal{F}$ of the crop dataset $\mathcal{C}$. The k-means clustering algorithm assigns one cluster id in [1, K] to each feature $\mathbf{f}_{j} \in \mathcal{F}$ and its corresponding segment $\mathbf{s^{X}_{i}} \in \mathbf{S^{X}}$, where $\mathbf{S^{X}} \in \mathcal{S}$.
\begin{figure}[ht]
\centering
\includegraphics[width=3.4in]{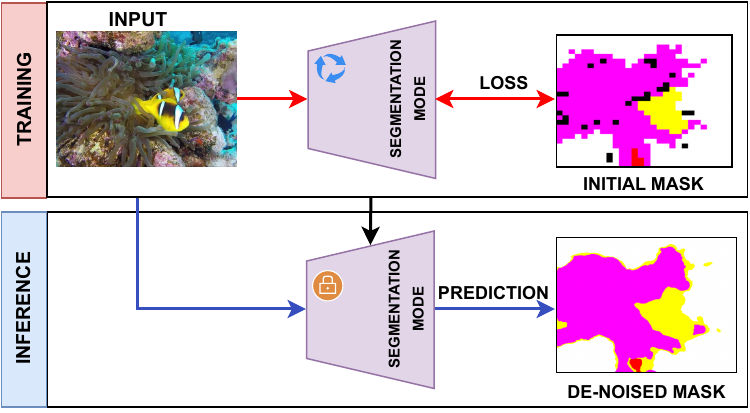}
	\caption{De-noising \textit{initial pseudo-annotated segmentation masks} from the second last step using deep learning-based segmentation model.}
	\label{fig:6}
\end{figure}

\begin{figure*}[!t]
\centering
\includegraphics[width=\textwidth]{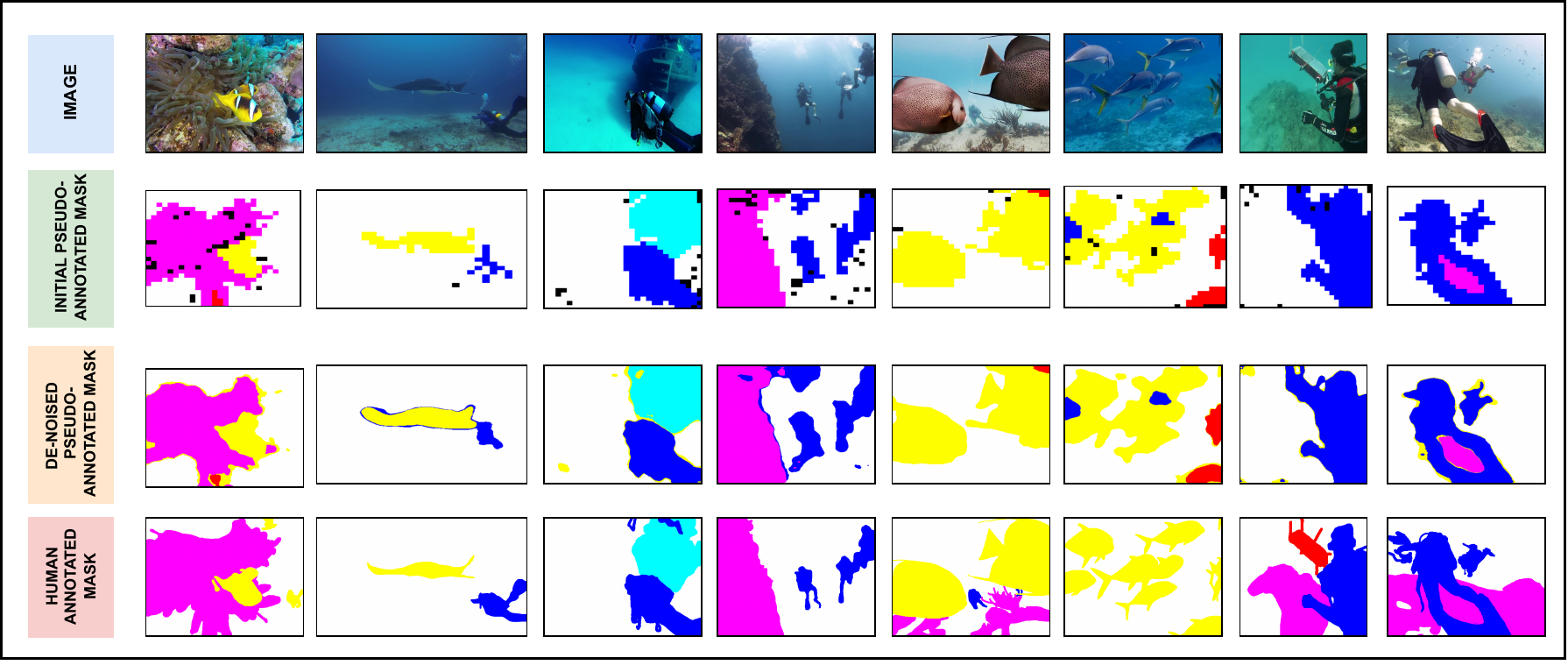}
	\caption{Visual comparison of \textit{Initial Pseudo-annotated Segmentation Masks}, Segment model's prediction, i.e., De-noised Pseudo-annotated Segmentation Masks, and Ground Truth Masks.}
	\label{fig:5}
\end{figure*}
\subsection{Create Initial Pseudo-annotated Masks}\label{subsecM5}
Next, to produce the pseudo-annotated segmented mask $\mathbf{M^{X}}$ of image $\mathbf{X}$, we assign the cluster id $\mathbf{k}$ of the corresponding segment $\mathbf{s^{X}_{i}}$ to the image pixel within this segment. The pixels belonging to the noise segment remain unlabeled in this step. As shown in the rightmost part of Figure \ref{fig:4}, we perform this operation on each image of the training dataset $\mathcal{X}$ to distill its corresponding Pseudo-annotated segmentation mask set $\mathcal{M}$. The second row of Figure \ref{fig:5} displays an exemplar subset of the \textit{initial pseudo-annotated segmentation masks} set $\mathcal{M}$ of the corresponding image from the first row. 
\subsection{Pseudo-mask De-noising and Smoothing}\label{subsecM6} 
An \textit{initial pseudo-annotated segmentation mask} generated in the last stage contains a few unlabeled patches and uneven object boundaries. As shown in Figure \ref{fig:6}, this step aims to generate a fully annotated smooth version of each \textit{initial pseudo-annotated segmentation mask}. We achieve this by harnessing the potential of deep learning. We utilize \textit{initial pseudo-annotated segmentation masks} $\mathcal{M}$ from the previous stage as pseudo-ground truth to train a segmentation model, i.e., DeepLabV3\cite{chen2017rethinking} from scratch (training part of Figure \ref{fig:6}). If most of the pixels in the pseudo-annotated segmentation mask belong to the same class, we drop it from the training dataset. The segmentation model computes the cross-entropy loss $\mathcal{L}^{T}$ between its prediction $\mathbf{Y}^{T} \in \mathbf{R}^{(K\times H \times W)}$ and pseudo-ground truth $\mathbf{\hat{Y}} \in \mathbf{R}^{(K\times H \times W)}$ of image $\mathbf{X}$ as given in equation \ref{e3}. The pseudo-ground truth $\mathbf{\hat{Y}} \in \mathbf{R}^{(K\times H \times W)}$ is one-hot encoded version of pseudo-annotated segmentation mask $\mathbf{M}^{X}$. The cross-entropy loss only computes loss for the labeled pixels. Hence, the unlabeled pixels belonging to noisy image segments get ignored in the training part. 
\begin{align} \label{e3}
    \mathcal{L}^{T} = 
    {
        \sum_{(H,W)} CE \Big(\mathbf{Y}^{T}_{(H,W)}, \mathbf{\hat{Y}}_{(H,W)} \Big)
    }
\end{align}
As shown in the inference part of Figure \ref{fig:6}, a trained segmentation model can predict the label for all pixels within an image. We extract the prediction of the segmentation model over the dataset's training and validation set for evaluation. The third row of Figure \ref{fig:5} displays the predictions of the segmentation model for the corresponding \textit{initial pseudo-annotated segmentation masks} from the second row. Algorithm 1 summarises the overall process of the proposed method, i.e., DatUS$^{2}$.
\section{EXPERIMENTS} \label{sec4}
\subsection{Datasets}
We justify the effectiveness of our model by experimenting with two popular segmentation datasets, SUIM \cite{islam2020semantic} and COCO-Stuff \cite{caesar2018coco}. The SUIM dataset is a collection of images from the underwater environment. The reason for choosing the underwater dataset is that the labeled dataset in the underwater environment is relatively limited. The COCO dataset is a large-scale scene-centric dataset with images from various domains. 

\textbf{SUIM:} It contains 1525 underwater train/validation images and 110 test images from underwater environments with human-annotated masks. The dataset recognizes eight visual classes, i.e., human divers, Aquatic plants/Sea-grass, Wrecks/Ruins, Robots (AU-Vs/ROVs/Instruments), Reefs/Invertebrates, Fish/Vertebrates, Sea-floor/Rocks, and Background. Following the previous work \cite{islam2020semantic}, we use the six-course visual categories by combining three classes: Aquatic-plants/Sea-grass, Sea-floor/Rocks, and Background.

\textbf{COCO:} It comprises a vast collection of scene-centric images, containing 80 categories for objects ('things') and 91 categories for background elements ('stuff'). We adopt the pre-processing from \cite{cho2021picie,ji2019invariant}, which forms coarse visual classes of 27 categories, i.e., 15 stuff and 12 things. 

\subsection{Setup}
Our proposed method, i.e., DatUS$^{2}$ works upon the ViT model \cite{dosovitskiy2020image} pre-trained with the original implementation of existing state-of-the-art self-supervised training schemes \cite{caron2021emerging, oquab2023dinov2, zhou2021ibot, zhou2022mugs, an2023unicom}. 
The proposed method, i.e., DatUS$^{2}$, utilizes the Louvain algorithm \cite{blondel2008fast} in the \textit{Discover Image Segments} step for hierarchical clustering of patch embeddings. Based on the qualitative observation, we consider the segments with more than five patches valid for pseudo labeling, i.e., the hyper-parameter $\tau = 5$, irrespective of the model and patch sizes. The proposed method, i.e., DatUS$^{2}$, utilizes the k-means clustering algorithm to assign the pseudo-labels for valid segments. We utilize the k-means clustering implementation for unsupervised image classification of the paper \cite{zheltonozhskii2020self} with necessary modification. In the last step of the proposed method, i.e., the \textit{Pseudo-mask De-noising and Smoothing step}, we utilize the supervised setup of the paper \cite{islam2020semantic} to train DeepLab \cite{chen2017rethinking} for the segmentation task of SUIM dataset. We use the \textit{initial pseudo-annotated segmentation masks} from the second last step of the proposed method, i.e., DatUS$^{2}$, as ground truth for training. Also, we test another version of the proposed method, i.e., DatUS$^{2}$ for ablation study by replacing the pre-trained ViT model with the CNN model pre-trained with MoCo \cite{he2020momentum} self-supervised training scheme in the \textit{Segment-wise Pseudo Labeling} step. We reproduce the feature extraction setup of \cite{zheltonozhskii2020self} to utilize the pre-trained feature extractor, i.e., Resnet-18 \cite{he2016deep} of the MoCo \cite{he2020momentum} framework. 
\begin{figure*}[h]
    \centering
    \subfloat[]{\includegraphics[width=2.3in]{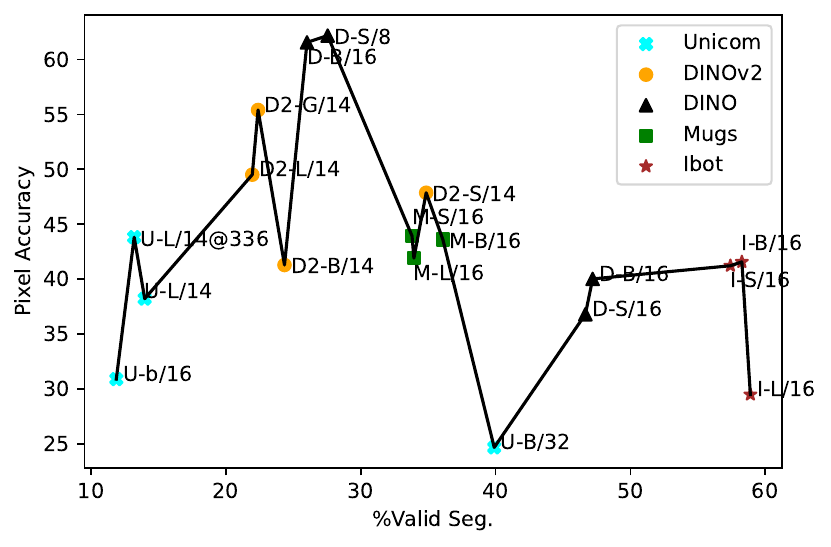}}
    \hfill
    \subfloat[]{\includegraphics[width=2.3in]{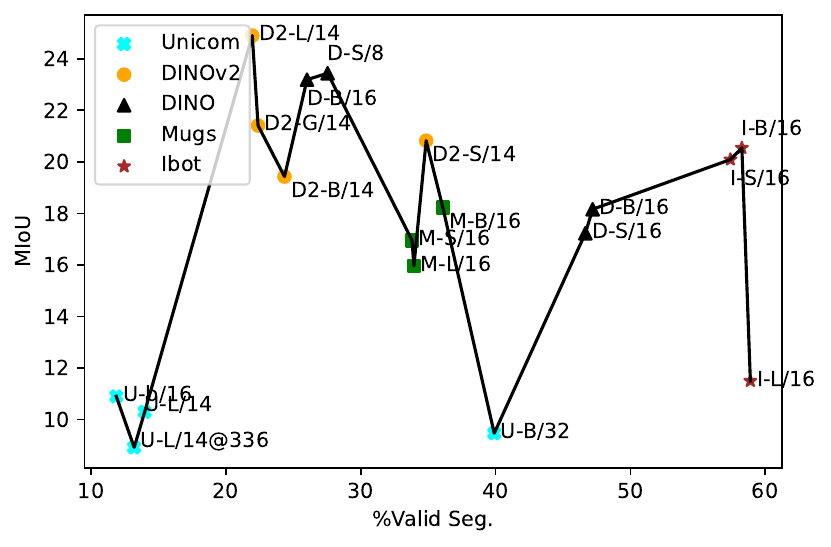}}
    \hfill
    \subfloat[]{\includegraphics[width=2.33in]{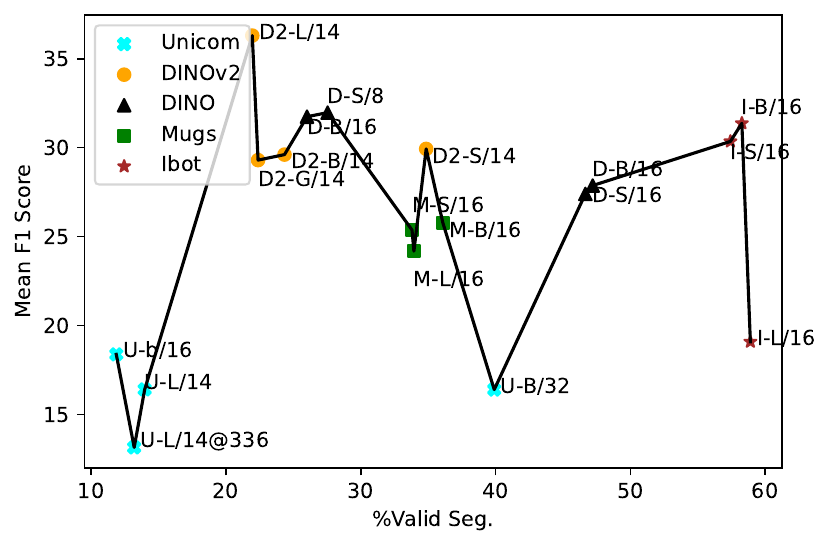}}
    \caption{The sub-figures (a),(b), and (c) plot Pixel Accuracy, MIoU, and Mean F1 Score form Table \ref{tab:table1} with the percentage of valid pixel discovered by the proposed method, i.e., DatUS$^{2}$ utilizing ViT models pre-trained with state-of-the-art self-supervised training schemes from Table \ref{tab:table2} (SUIM dataset).The U, D2, D, M, and I in plot Unicom, DINOv2, DINO, Mugs, and IBoT self-supervised training scheme, respectively. The S, B, L, and G denote the ViT model size in ascending order. The 8, 14, 16, and 32 denote the patch sizes accepted by the respective model.}
    \label{fig:7}
\end{figure*}

\subsection{Evaluation Protocol and Matrices}
We evaluate our proposed method in three steps: (1) First, we explore the efficacy of the proposed method, i.e., DatUS$^{2}$, as a downstream task for self-supervised training schemes on the training set in subsection \ref{Downstream}, (2) Additionally, we evaluate the \textit{Pseudo-mask De-noising and Smoothing} step separately in subsection \ref{Denoise}, and (3) Finally, we evaluate the proposed model for the semantic segmentation task on validation test, and compare with existing state-of-the-art unsupervised semantic segmentation methods in subsection \ref{SOTA}. For the first and third evaluation steps, we compare the \textit{initial pseudo-annotated segmentation masks} of the proposed method, i.e., DatUS$^{2}$, with ground truth masks of both the SUIM and COCO datasets. In the second evaluation step, we compare the de-noised masks with corresponding \textit{initial pseudo-annotated segmentation masks} on the training and validation set of the SUIM dataset.


The pseudo-annotated segmentation masks have pixel labels belonging to clusters 1 to K. To compare the pseudo-segmentation masks with ground truth in the unsupervised setup, we use Hungarian matching \cite{kuhn2005hungarian} for mapping the K number of predicted cluster labels with the C number of the ground truth classes and $K \geq C$. Next, we create the confusion matrix based on the mapping obtained from the Hungarian matching. Following the previous work \cite{vobecky2022drive}, we compute MIoU and pixel accuracy from the confusion matrix and consider $K-C$ un-mapped classes as false negatives (if applicable). In the case of \textit{initial pseudo-annotated segmentation masks}, we drop the unlabeled pixels belonging to noisy segments for evaluation. Unless stated, we consider $K = C$ for the evaluation process. In the ablation study, we analyze the sensitivity of DatUS$^{2}$ for different numbers of clusters, i.e., $K \geq C$. We use three popular matrices of the semantic segmentation task, i.e., Mean Intersection over Union (MIoU), Pixel Accuracy (PAcc.), and Dice Coefficient or F1-score (Avg. F1). The Intersection over Union (IoU) computes the overlap region of the predicted region and ground truth for a class, and MIoU averages the result overall categories. The pixel accuracy calculates the fraction of correctly predicted pixels among all pixels. The Dice Coefficient or F1-score is the harmonic mean of the precision and recall and a popular performance metric for class imbalanced data. 
\subsection{Comparative Analysis of Recent Self-supervised Training Schemes with Proposed Method, i.e., DatUS$^{2}$ on Training Set} \label{Downstream}
In this subsection, we evaluate the recent self-supervised training schemes' capacity to learn quality semantic properties at patch-level embeddings based on the performance of our proposed downstream task, i.e., DatUS$^{2}$. We consider various current state-of-the-art self-supervised training schemes like DINO, IBoT, Mugs, DINOv2, and Unicom (Uni) for the experiment. 
For the SUIM dataset, we test multiple versions of pre-trained ViT models (based on the model and patch sizes) to find the best-performing version corresponding to each self-supervised training scheme. Next, we consider the best-performing ViT model of each self-supervised training scheme to experiment on the large-scale dataset, i.e., the COCO dataset.

\begin{table}[ht]
\caption{Comparative analysis of recent self-supervised training schemes for semantic segmentation on training set of SUIM dataset based on the performance of our proposed downstream task DatUS$^{2}$. We underline the best value in each column within a self-supervised training scheme. The top three values in each column among all self-supervised schemes are color-coded with red, blue, and magenta, respectively. \label{tab:table1}}
\centering
\begin{tabular}{@{\hspace{0.1cm}}l@{\hspace{0.1cm}}c@{\hspace{0.1cm}}c@{\hspace{0.1cm}}c@{\hspace{0.1cm}}c@{\hspace{0.1cm}}c}
\hline
\multirow{2}{*}{\textbf{Method}} & \multirow{2}{*}{\textbf{Params.}} & \multirow{2}{*}{\textbf{\begin{tabular}[c]{@{}l@{}}Pre-training\\ Dataset\end{tabular}}} & \multirow{2}{*}{\textbf{MIoU}} & \multirow{2}{*}{\textbf{PAcc.}} & \multirow{2}{*}{\textbf{Avg. F1}}\\
&&&&&\\
\hline
DatUS$^{2}$(DINO-S/8) & 21M & ImageNet-1K & \textcolor{magenta}{\textbf{23.19}} & \textbf{\textcolor{blue}{61.56}} & \textcolor{magenta}{\textbf{31.74}} \\
DatUS$^{2}$(DINO-S/16) & 21M & ImageNet-1K & 17.23 & 36.79 & 27.40 \\
DatUS$^{2}$(DINO-B/8) & 85M & ImageNet-1K & \textbf{\textcolor{blue}{\uline{23.44}}} & \textbf{\textcolor{red}{\uline{62.18}}} & \textbf{\textcolor{blue}{\uline{31.97}}} \\
DatUS$^{2}$(DINO-B/16) & 85M & ImageNet-1K & 18.16 & 40.01 & 27.87 \\
\hline
DatUS$^{2}$(IBoT-S/16) & 21M & ImageNet-1K & 20.09 & 41.21 & 30.35 \\
DatUS$^{2}$(IBoT-B/16) & 85M & ImageNet-1K & \uline{20.54} & \uline{41.56} & \uline{31.37} \\
DatUS$^{2}$(IBoT-L/16) & 307M & ImageNet-1K & 11.50 & 29.48 & 19.09 \\
\hline
DatUS$^{2}$(Mugs-S/16) & 21M & ImageNet-1K & 16.99 & \uline{43.93} & 25.37 \\
DatUS$^{2}$(Mugs-B/16) & 85M & ImageNet-1K & \uline{18.22} & 43.59 & \uline{27.77} \\
DatUS$^{2}$(Mugs-L/16) & 307M & ImageNet-1K & 15.96 & 41.90 & 24.18 \\
\hline
DatUS$^{2}$(DINOv2-S/14) &  21M & LVD-142M & 20.82 & 47.86 & 29.93 \\
DatUS$^{2}$(DINOv2-B/14) & 86M & LVD-142M & 19.43 & 41.28 & 29.61 \\
DatUS$^{2}$(DINOv2-L/14) & 300M & LVD-142M & \textbf{\textcolor{red}{\uline{24.90}}} & 49.51 & \textbf{\textcolor{red}{\uline{36.3}}} \\
DatUS$^{2}$(DINOv2-G/14) & 1100M & LVD-142M & 21.41 & \textcolor{magenta}{\textbf{\uline{55.40}}} & 29.30 \\
\hline
DatUS$^{2}$(Uni-B/16) & 21M & LAION 400M & \uline{10.90} & 30.88 & \uline{18.38} \\
DatUS$^{2}$(Uni-B/32) & 21M & LAION 400M & 09.47 & 24.66 & 16.40 \\
DatUS$^{2}$(Uni-L/14) & 85M & LAION 400M & 10.30 & \uline{38.21} & 16.41 \\
DatUS$^{2}$(Uni-L/14@336) & 85M & LAION 400M & 08.92 & 43.81 & 13.15 \\
\hline
\end{tabular}
\end{table}

\begin{table}[ht]
\caption{Comparing the total number of segments and percentage of valid segments discovered by the proposed method, i.e., DatUS$^{2}$ utilizing different state-of-the-art self-supervised training schemes on the SUIM dataset. \label{tab:table2}}
\centering
\begin{tabular}{l@{\hspace{1.2cm}}c@{\hspace{1.2cm}}c}
\hline
\textbf{Method} & \multirow{1}{*}{\textbf{\#Total Seg}} & \multirow{1}{*}{\textbf{\%Valid Seg.}}\\ \hline
 
DatUS$^{2}$(Uni-B/16) & 50,912 & 11.86 \\
DatUS$^{2}$(Uni-L/14@336)& 59,017 & 13.19 \\
DatUS$^{2}$(Uni-L/14) & 36,623 & 13.97 \\
DatUS$^{2}$(DINOv2-L/14) & 34,633 & 21.95 \\
DatUS$^{2}$(DINOv2-G/14) & 32,254 & 22.38  \\
DatUS$^{2}$(DINOv2-B/14) & 26,665 & 24.34  \\
DatUS$^{2}$(DINO-S/8) & 26,273 & 26.01  \\
DatUS$^{2}$(DINO-B/8) & 26,226 & 27.54  \\
DatUS$^{2}$(Mugs-S/16) & 24,125 &  33.80  \\
DatUS$^{2}$(Mugs-L/16) & 24,498 & 33.94 \\
DatUS$^{2}$(DINOv2-S/14) & 16,580 & 34.85  \\
DatUS$^{2}$(Mugs-B/16) & 21,310 & 36.09 \\
DatUS$^{2}$(Uni-B/32)  & 11,136 & 39.90  \\
DatUS$^{2}$(DINO-S/16) & 9,485 & 46.66  \\
DatUS$^{2}$(DINO-B/16) & 9,552 & 47.20  \\
DatUS$^{2}$(IBoT-S/16) & 9,155 & 57.42  \\
DatUS$^{2}$(IBoT-B/16) & 9,259 & 58.27  \\
DatUS$^{2}$(IBoT-L/16) & 9,187 & 58.9  \\
\hline
\end{tabular}
\end{table}
\textbf{SUIM:} 
In Tables \ref{tab:table1} and \ref{tab:table2}, we compare the performance of our proposed method, i.e., DatUS$^{2}$ utilizing multiple versions of the pre-trained ViT model for existing self-supervised training schemes. Table \ref{tab:table1} compares their ability to capture semantic information during pre-training with three performance matrices, i.e., MIoU, Pixel Accuracy (PAcc.), and Average F1 score (Avg. F1). Table \ref{tab:table2} lists the total number of segments (\#Total Seg.) and percentage of valid segments (\% Valid Seg.) discovered by the proposed method, i.e., DatUS$^{2}$ in different settings. Figure \ref{fig:7}a, \ref{fig:7}b, and \ref{fig:7}c show the influence of the percentage of valid segments (\% Valid Seg.) on the pixel accuracy, MIoU, and average F1 Score matrices of initial pseudo-annotated segmented masks, respectively.  

Each valid segment of an image discovered by our proposed method, i.e., DatUS$^{2}$, with an underlying self-supervised training scheme captures an object from the scene. Also, each segment is a collection of meaningful patches. Figure \ref{fig:7} shows that the self-supervised training schemes with vision transformer of the large-size patch, i.e., 16 and 32, perform poorly for all matrices. The self-supervised training schemes with a vision transformer of large size patch end up decomposing scenes in fewer segments consisting of one or more objects with uneven boundaries. The information presented in Table \ref{tab:table2} confirms a reduction in the total number of segments as patch sizes increase. In contrast, the best-performing self-supervised training schemes have a vision transformer of the smallest patch size, i.e., the segment made of smaller patches decomposes scenes into objects having more accurate object boundaries. 

In summary, Table \ref{tab:table1}, Table \ref{tab:table2}, and Figure \ref{fig:7} conclude that the performance of the proposed method, i.e., DatUS$^{2}$, increases as the patch size decreases. Also, the best-performing models, DINO-B/8, DINO-S/8, DINOv2-L/14, and DINOv2-S/14, discover around 25-30 percent of valid segments. 
Among all the ViT models pre-trained with existing state-of-the-art self-supervised training schemes, DINO-B/8 achieves the highest pixel accuracy (62.18), with the second highest MIoU (23.44) and Mean F1 Score (31.97). Also, DINOv2-L/14 achieves high MIoU (24.90) and Mean F1-score (36.3), with the third highest pixel accuracy (49.51). Table \ref{tab:table1} shows that DINO-b/8 (23.44 MIoU, 62.18 PA, 31.97 MF1), DINOv2-G/14  (21.41 MIoU, 55.40 PA, 29.30 MF1), IBoT-B/16 (20.54 MIoU, 41.56 PA, 31.37 MF1), Mugs-B/16  (18.22 MIoU, 43.59 PA, 27.77 MF1), Unicom-L/14@336 (08.92 MIoU, 43.81 PA, 13.15 MF1) are best-performing ViT models of the DINO, DINOv2, IBoT, Mugs, and Unicom Scheme, respectively. The DINO-based schemes dominate all other schemes because they support the first two smallest patch sizes, i.e., 8 and 14. Also, It utilizes the contrastive learning-based object, which is known to generalize well across images and tasks. 
\begin{table*}[ht]
\caption{Comparative analysis of best performing ViT model for each self-supervised training scheme from Table \ref{tab:table1} on the training set of COCO dataset based on the performance of our proposed downstream task DatUS$^{2}$. The top three values in each column among all self-supervised schemes are color-coded with red, blue, and magenta, respectively. \label{tab:table1.1}}
\centering
\begin{tabular}{lcccccccc}
\hline
\multirow{2}{*}{\textbf{Method}} & \multirow{2}{*}{\textbf{Params.}} & \multirow{2}{*}{\textbf{\begin{tabular}[c]{@{}l@{}}Pre-training\\ Dataset\end{tabular}}} & \multicolumn{3}{c}{\textbf{Semantic Segmentation}} && \multicolumn{2}{c}{\textbf{Semant Discovery}} \\
\cline{4-6}\cline{8-9}
& & & \textbf{MIoU} & \textbf{PAcc.} & \textbf{Avg. F1}&&\textbf{\#Total Seg.}&\textbf{\%Valid Seg.}\\
\hline

DatUS$^{2}$(DINO-B/8) & 85M & ImageNet-1K &\textbf{\textcolor{red}{10.60}}&\textbf{\textcolor{red}{31.50}}&\textbf{\textcolor{red}{16.77}}&&2,65,048&25.90\\

DatUS$^{2}$(IBoT-B/16) & 85M & ImageNet-1K & \textbf{\textcolor{blue}{09.86}}&\textbf{\textcolor{blue}{25.90}}&\textbf{\textcolor{blue}{16.22}}&&1,88,153&70.48\\
DatUS$^{2}$(Mugs-B/16) & 85M & ImageNet-1K &\textcolor{magenta}{09.03}&\textcolor{magenta}{22.26}&\textcolor{magenta}{14.96}&&2,99,211&40.03\\
DatUS$^{2}$(DINOv2-G/14) & 1100M & LVD-142M &08.44&23.03&14.38&&2,38,954&24.07\\
\hline
\end{tabular}
\end{table*}

\begin{figure}[ht]
\centering
\includegraphics[width=3.4in]{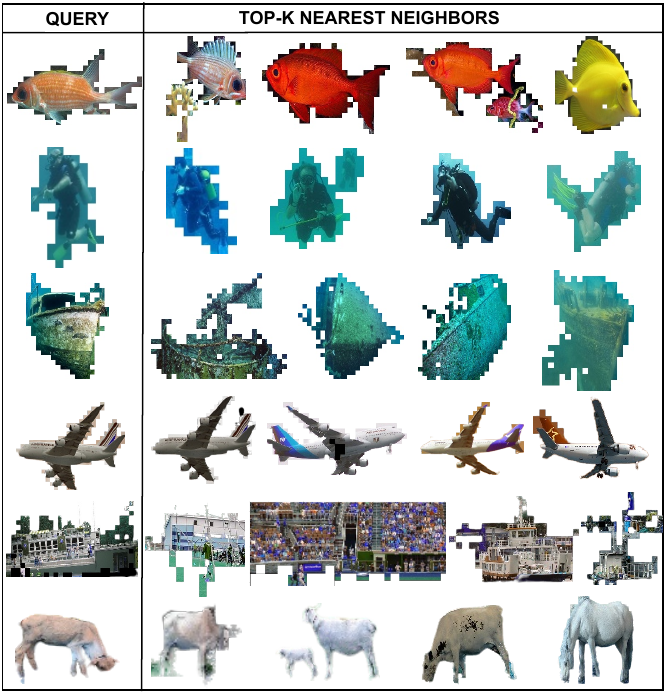}
	\caption{The K-nearest neighbor-based segment retrieval over valid segments discovered in the intermediate step of the proposed method, i.e., DatUS$^{2}$.}
	\label{fig:9}
\end{figure}

\textbf{COCO:}  
Now, we utilize the best-performing ViT model of each self-supervised scheme to evaluate, i.e., DatUS$^{2}$ performance on a training split of the large-scale dataset, i.e., the COCO dataset. Table \ref{tab:table1.1} presents the performance of the proposed method, i.e., DatUS$^{2}$ on the training set of the COCO dataset. Similar to the SUIM dataset, the performance of our proposed method with DINO, i.e., DatUS$^{2}$(DINO-B/8), is better than all existing self-supervised schemes for the COCO dataset. Additionally, Figure \ref{fig:9} shows the correlation between the segments discovered by the unsupervised graph clustering algorithm in the intermediate step of the proposed method, i.e., DatUS$^{2}$ utilizing the best-performing model, i.e., DINO-B/8. We perform the segment retrieval using a k-nearest neighbor algorithm over the valid segments' feature vectors (CLS token) for qualitative evaluation. Figure \ref{fig:9} illustrates that the proposed method, i.e., DatUS$^{2}$, decomposes the scenes into meaningful correlated segments, which is crucial for further steps of segmentation.

In summary, the performance of the DatUS$^{2}$ depends upon the quality of valid image segments discovered by the unsupervised graph clustering algorithm and the quality of visual groups of valid segments identified by k-means clustering in intermediate steps. The k-means clustering used in the \textit{Segment-wise Pseudo Labeling} step suffers from the curse of dimensionality. Although DatUS$^{2}$ discover highly correlated image segments, the \textit{Segment-wise Pseudo Labeling} step dictates the performance on segmentation matrices. Also, Table \ref{tab:table1} and \ref{tab:table1.1} show that the proposed method favors a smaller dataset due to the limitation of k-means clustering.
\subsection{Evaluation of Pseudo-mask De-noising and Smoothing}\label{Denoise}
In this sub-section, we evaluate the additional step of the proposed method, i.e., \textit{Pseudo-mask De-noising and Smoothing}. Table \ref{tab:table5} compares the de-noised masks with DatUS$^{2}$'s output, i.e., \textit{initial pseudo-annotated segmentation masks} on the training and validation set. The de-noised masks are the prediction of the segment model trained in \textit{Pseudo-mask De-noising and Smoothing} step. The segmentation model is trained with image and corresponding \textit{initial pseudo-annotated segmentation masks} (DatUS$^{2}$'s output) on the training set. To compute the evaluation matrices, we evaluate each set of masks with their corresponding ground truth masks.

The performance of initial and de-noised masks on the training set shows that the segmentation model makes meaningful predictions for the unlabeled pixels, i.e., pixels belonging to noisy segments of the DatUS$^{2}$'s \textit{initial pseudo-annotated segmentation masks}. The second and third rows of Figure \ref{fig:5} visually compare the initial pseudo-annotated mask and their corresponding predictions. In addition to that, the de-noised prediction of the segmentation model on the unseen images from the validation set of the SUIM dataset achieves almost similar accuracy. It validates that the initial pseudo-segmentation masks possess significant semantic and spatial characteristics for supervising a deep learning model.

\begin{table}[ht]
\caption{Comparing the performance of DatUS$^{2}$(DINO-B/8)'s \textit{initial pseudo-annotated segmentation masks} on training set, segment model's prediction on train and validation set of SUIM dataset.}\label{tab:table5}
\centering
\begin{tabular}{lcccc}
\hline
 \textbf{Mask Type} &  \textbf{Dataset split} & \multicolumn{1}{c}{\textbf{MIoU}} &  \multicolumn{1}{c}{\textbf{PAcc.}} &  \multicolumn{1}{c}{\textbf{Avg. F1}} \\ \hline

\multirow{2}{*}{Initial Masks} & Train & 23.19  & 61.56  & 31.74\\ \cline{2-5}
& Val & 28.48 & 64.67  & 39.52\\ \hline
\multirow{2}{*}{De-noised Masks} & Train & 24.19 & 61.43  & 31.34\\ \cline{2-5}
  & Val & 23.21 & 61.82  & 32.18\\ \hline
\end{tabular}
\end{table}
\subsection{Comparative Analysis of DatUS$^{2}$ with Existing State-of-the-art (SOTA) Unsupervised Dense Semantic Segmentation method}\label{SOTA}
The results from the previous sub-subsections show that the DINO self-supervised training scheme with ViT-B/8, i.e., DINO-B/8 learns high-quality patch embeddings for our proposed downstream task, i.e., DatUS$^{2}$ for both datasets. Hence, we compare the performance of DatUS$^{2}$ with the existing SOTA method for unsupervised semantic segmentation, i.e., STEGO \cite{hamilton2022unsupervised}, on the validation set of the SUIM and COCO datasets. To compute STEGO results on SUIM dataset, we train multiple ViT models (ViT-s/8 and ViT-B/8) from scratch.
Table \ref{SOTA} shows that the proposed method DatUS$^{2}$ (ViT-B/8) outperforms the STEGO (ViT-S/8) on the SUIM dataset with 15.02 \% MiOU and 21.47 \% Pixel accuracy. The performance further improves up to 37.40 \% MiOU and 31.44 \% with the overclustering. Also, the proposed method achieves a competitive level of accuracy on the large-scale dataset, i.e., COCO dataset.
The performance of the proposed downstream task DatUS$^{2}$ depends upon the quality of patch-level feature representation learned during self-supervised training. The experimental results reveal that there is a scope to improve the self-supervised learning schemes to learn informative patch embeddings for a complex and scene-centric dataset like the COCO dataset.

\begin{table}[ht]
\caption{Comparison with state-of-the-art unsupervised semantic segmentation methods on the validation set. The top two values in each column among all self-supervised schemes are color-coded with red and blue, respectively.}\label{tab:SOTA}
\centering
\begin{tabular}{lccc}
\hline
\multicolumn{4}{c}{\multirow{2}{*}{\textbf{SUIM Dataset}}} \\
&&&\\
\hline
\textbf{Method} & \textbf{Backbone} &\textbf{MIoU} & \textbf{PAcc.} \\
\hline
STEGO \cite{hamilton2022unsupervised} & ViT-S/8 & 24.76 & 53.24 \\
STEGO \cite{hamilton2022unsupervised} & ViT-B/8 & 21.51 & 45.42 \\
DatUS$^{2}$ (Our) & ViT-S/8 &  23.19 & 61.56 \\
DatUS$^{2}$ (Our) & ViT-B/8 &  \textcolor{blue}{\textbf{28.48}} & \textcolor{blue}{\textbf{64.67}} \\
DatUS$^{2}$+Overclust.(K=7) (Our) & ViT-B/8 & \textcolor{red}{\textbf{34.02}}  & \textcolor{red}{\textbf{69.98}}\\
\hline
\hline
\multicolumn{4}{c}{\multirow{2}{*}{\textbf{COCO Dataset}}}\\
&&&\\
\hline
\textbf{Method} & \textbf{Backbone} &\textbf{MIoU} & \textbf{PAcc.} \\
\hline
IIC \cite{ji2019invariant} & ResNet/VGG11& 6.7 & 21.8\\
PiCIE \cite{cho2021picie} & FPN+ResNet18 & 14.36 & \textcolor{blue}{\textbf{49.99}} \\
STEGO \cite{hamilton2022unsupervised} & ViT-B/8 & \textcolor{red}{\textbf{28.02}} & \textcolor{red}{\textbf{56.45}} \\
DatUS$^{2}$ (Our) & ViT-B/8 &  12.01 &  34.11 \\
DatUS$^{2}$+Overclust.(K=63) (Our) & ViT-B/8 & \textcolor{blue}{\textbf{15.97}}  & 33.94\\
\hline
\hline
\end{tabular}
\end{table}

\subsection{Ablation Study}
This subsection explores possible architectural and theoretical changes to improve the performance of the proposed method for unsupervised semantic segmentation tasks. We continue the experiment from the subsection \ref{Downstream} on the training set of the SUIM dataset without labels for further study. 

\textbf{Segment-wise feature extractor:} One possible modification is to replace the vision transformer-based feature extractor in the \textit{Segment-wise Pseudo Labeling} step of the proposed method, i.e., DatUS$^{2}$. We analyze the proposed method's performance with the self-supervised CNN-based feature extractor for all the self-supervised ViT models from Table \ref{tab:table1}. Table \ref{tab:table3} presents a comparative analysis of two different versions of the proposed method, i.e., with self-supervised vision transformer-based and CNN-based feature extractors for all three performance matrices. As shown in Table \ref{tab:table1}, the top-performing models come from the DINO and DINOv2 self-supervised training schemes. Table \ref{tab:table3} shows that in top-performing models, the vision-transformer-based feature extractor dominates the CNN-based feature extractor. Unlike that, with average-performing models, CNN-based feature extractors surpass the performance of vision transformer-based feature extractors.
\begin{table}[ht]
\caption{Ablation study of the proposed method, i.e., DatUS$^{2}$ with self-supervised vision transformer-based and CNN-based feature extractor on SUIM dataset.\label{tab:table3}}
\centering
\begin{tabular}{l@{\hspace{0.2cm}}c@{\hspace{0.1cm}}c@{\hspace{0.1cm}}c@{\hspace{0.1cm}}c@{\hspace{0.1cm}}c@{\hspace{0.1cm}}c@{\hspace{0.1cm}}c@{\hspace{0.1cm}}c}
\hline
\multirow{2}{*}{\textbf{Method}} & \multicolumn{2}{c}{\textbf{MIoU}} & &\multicolumn{2}{c}{\textbf{PAcc.}} & & \multicolumn{2}{c}{\textbf{Avg F1}}\\ \cline{2-3} \cline{5-6} \cline{8-9}
    & \textbf{Trans.} & \textbf{CNN} && \textbf{Trans.} & \textbf{CNN} && \textbf{Trans.} & \textbf{CNN}\\
\hline
DatUS$^{2}$(DINO-S/8) & \textbf{23.19} & 19.37 && \textbf{61.56} & 48.27 && \textbf{31.74} & 29.03\\
DatUS$^{2}$(DINO-S/16) & 17.23 & \textbf{20.68} && 36.79 & \textbf{47.85} && 27.40 & \textbf{31.12}\\
DatUS$^{2}$(DINO-B/8) & \textbf{23.44} & 20.33 && \textbf{62.18} & 50.38 && \textbf{31.97} & 30.00\\
DatUS$^{2}$(DINO-B/16) & 18.16 & \textbf{22.10} && 40.01 & \textbf{50.27} && 27.87 & \textbf{32.91}\\
\hline

DatUS$^{2}$(IBoT-S/16) & 20.09 & \textbf{22.59} &&41.21 & \textbf{47.55} && 30.35 & \textbf{33.57}\\
DatUS$^{2}$(IBoT-B/16) & 20.54 & \textbf{21.57} && 41.56 & \textbf{47.51} && 31.37 & \textbf{32.48}\\
DatUS$^{2}$(IBoT-L/16) & 11.50 & \textbf{19.81} && 29.48 & \textbf{47.55} && 19.09 & \textbf{29.56}\\
\hline

DatUS$^{2}$(Mugs-S/16) & \textbf{16.99} & 16.96 && \textbf{43.93} & 38.97 && 25.37 & \textbf{26.81}\\
DatUS$^{2}$(Mugs-B/16) & \textbf{18.22} & 16.40 && \textbf{43.59} & 40.32 && \textbf{27.77} &  25.37\\
DatUS$^{2}$(Mugs-L/16) & \textbf{15.96} & 14.13 && \textbf{41.90} & 33.89 && \textbf{24.18} &  22.61\\
\hline

DatUS$^{2}$(DINOv2-S/14) & 20.82 & \textbf{21.82} && 47.86 & \textbf{49.55}  && 29.93 & \textbf{32.41}\\
DatUS$^{2}$(DINOv2-B/14) & \textbf{19.43} & 16.02 && 41.28 & \textbf{41.31}&& \textbf{29.61} & 24.36\\
DatUS$^{2}$(DINOv2-L/14) & \textbf{24.90} & 16.99 && \textbf{49.51} & 41.92 && \textbf{36.3} & 26.13\\
DatUS$^{2}$(DINOv2-G/14) & \textbf{21.41} & 15.25 && \textbf{55.40} & 37.48 && \textbf{29.30} & 23.10\\
\hline

DatUS$^{2}$(Uni-B/16) & 10.90 & \textbf{12.70} && 30.88 & \textbf{34.16} && 18.38 & \textbf{20.58}\\
DatUS$^{2}$(Uni-B/32) & 09.47 & \textbf{12.95} && 24.66 & \textbf{32.94} && 16.40 & \textbf{21.14}\\
DatUS$^{2}$(Uni-L/14) & 10.30 & \textbf{11.29} && \textbf{38.21} & 32.26 && 16.41 & \textbf{18.38}\\
DatUS$^{2}$(Uni-L/14@336) & \multirow{1}{*}{08.92} & \multirow{1}{*}{\textbf{11.60}} && \multirow{1}{*}{\textbf{43.81}} & \multirow{1}{*}{34.82} && \multirow{1}{*}{13.15} & \multirow{1}{*}{\textbf{18.40}}\\

\hline
\end{tabular}
\end{table}

\begin{figure}[ht]
\centering
\includegraphics[scale=0.4]{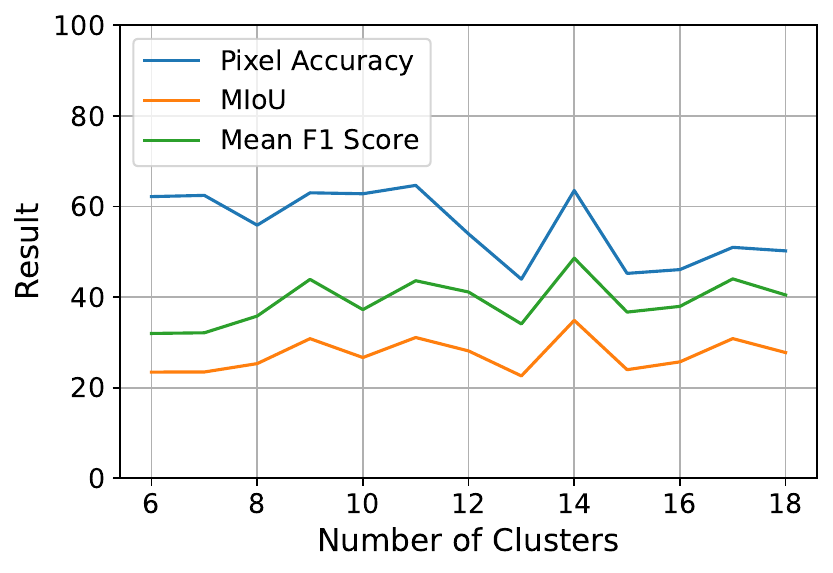}
	\caption{The impact of the number of clusters (K) for k-means in segment-wise pseudo-labeling step on the performance of the best-performing version of proposed method from Table \ref{tab:table1}, i.e., DatUS$^{2}$(DINO-B/8).} 
	\label{fig:8}
\end{figure}

\textbf{Overclustering:} In Table \ref{tab:table4}, we study the impact of the number of clusters (K) for k-means in segment-wise pseudo-labeling step on the performance of the best-performing model from Table \ref{tab:table1}, i.e., DINO-B/8. It shows that the proposed method achieves superior performance with overclustering.
Yet, Figure \ref{fig:8} illustrates that overclustering enhances the performance of the proposed method, i.e., DatUS$^{2}$ only up to a certain threshold, i.e., $k = 14$. Also, the overclustering impacts MIoU and Mean F1 Score more in contrast with the pixel accuracy of the \textit{initial pseudo-annotated segmentation masks}.
\begin{table}[ht]
\caption{Ablation study of the proposed method, i.e., DatUS$^{2}$(DINO-B/8) with a different number of pseudo-clusters on the SUIM dataset.\label{tab:table4}}
\centering
\begin{tabular}{c@{\hspace{1.2cm}}c@{\hspace{1.2cm}}c@{\hspace{1.2cm}}c}
\hline
 \textbf{\#Clusters(K)} & \textbf{MIoU} & \textbf{PAcc.} & \textbf{Avg. F1}\\
\hline
K=6& 23.44 & 62.18 &  31.97\\
K=7 & 23.46 & 62.45 & 32.12\\
K=8 & 25.31 & 55.88 & 35.81\\
K=9 & 30.82 & 63.02 & 43.91\\
  
K=10 & 26.65 & 62.81 & 37.23\\
K=11 & \textcolor{blue}{\textbf{31.08}} & \textcolor{red}{\textbf{64.66}} & 43.61\\
K=12 & 28.12 & 53.94 & 41.11\\
 
K=13 & 22.60 & 43.95 & 34.07\\
K=14 & \textbf{\textcolor{red}{34.87}} & \textcolor{blue}{\textbf{63.54}} & \textcolor{red}{\textbf{48.60}}\\
K=15 & 23.97 & 45.23 & 36.68\\
 
K=16 & 25.70 & 46.07 & 37.95\\
 
K=17 & 30.84 & 50.98 & \textbf{\textcolor{blue}{44.02}}\\
K=18 & 27.75 &  50.20 & 40.47\\
\hline
\end{tabular}
\end{table}
\section{Discussion and Future Work} \label{sec5}
Unlike the existing works, we propose a novel data-driven unsupervised semantic segmentation (DatUS$^{2}$) based on the self-supervised feature representation, i.e., patch embedding of the vision transformer. The work aims to establish a novel downstream task to evaluate the self-supervised training schemes of the vision transformer based on the semantic properties learned during training. Now, we present a range of concerns and potential avenues for future exploration. The initial query is whether the proposed method can be improved to mine a more accurate semantic mask like the dedicated unsupervised semantic segmentation training. The possible direction to explore is to reduce the ViT model's patch size further and introduce semantic-based information into self-supervised training like \cite{ouali2020autoregressive} to minimize the number of noisy segments. Second, it is unclear why some of the best-performing self-supervised training schemes for existing unsupervised computer vision tasks perform poorly for data-driven unsupervised semantic segmentation (DatUS$^{2}$). Third, we observed that the model performance for segmentation matrices is sensitive to the size and complexity of the dataset. The segment-wise pseudo-labeling step of the proposed method can be improved to mitigate this issue. The possible avenues for exploration are to fine-tune the self-supervised feature extractors with segment-wise crops and utilize different feature clustering methods like \cite{sarfraz2019efficient} to overcome the curse of dimensionality of k-means clustering.

In addition, the intermediate outcomes of the proposed approach, like image segments and the initial/final pseudo-segmentation masks, consist of high-quality semantic information that can be used as visual priors to boost the performance of various supervised and unsupervised computer vision applications. The proposed method is a promising approach to mining high-quality visual priors for various unsupervised computer vision tasks like image registration, end-to-end semantic segmentation, and key-point detection. Also, the model could be further optimized for various unsupervised computer vision applications like medical imaging, robotic vision, and autonomous driving, where semantic segmentation plays a crucial role.  

\section{Conclusion} \label{sec6}
In this paper, we study the applicability of self-supervised representation learning for dense unsupervised semantic segmentation with a vision transformer. We proposed a novel data-driven unsupervised semantic segmentation, i.e., DatUS$^{2}$, as a downstream task to analyze the effectiveness of recent self-supervised training schemes to learn a good quality patch-level feature representation. Additionally, we present a way to enhance the quality of an initial pseudo-annotated segmented mask by leveraging a deep learning-based segmentation model. These improved masks can serve as valuable visual priors for various computer vision applications. Also, we compared the best-performing version of the proposed method with the state-of-the-art methods for dense semantic segmentation for multiple datasets. The study reveals that self-supervised patch-level feature representations hold informative properties for pixel-based downstream tasks like dense semantic segmentation. Also, there is a scope to improve the existing self-supervised training schemes to learn more informative and discriminative patch-level representation for more complex scenes.

Finally, we perform an ablation study to improve further the proposed method's performance for the semantic segmentation task. We also present a range of concerns and potential avenues for future exploration. According to our knowledge, this is the first attempt to propose a dense semantic segmentation as a downstream task for self-supervised training schemes for a vision transformer. Our proposed method shows a new research direction to propose similar novel unsupervised downstream tasks like unsupervised depth estimation for a self-supervised training scheme. Such contributions can help develop a robust and generalized self-supervised training scheme applicable across the various supervised or unsupervised downstream tasks.




\bibliographystyle{IEEEtran}
\bibliography{main}
\vspace{-30pt}
\begin{IEEEbiography}[{\includegraphics[width=1in,height=1.25in,clip,keepaspectratio]{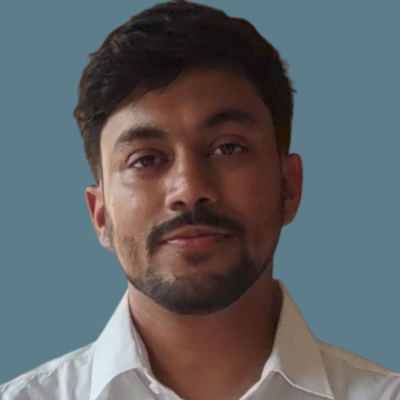}}]{Sonal Kumar} received the M.Tech. degree in Computer Science and Engineering from the National Institute of Technology Patna, and a B.E. degree in Information Science and Engineering from VTU University. Currently, he is actively engaged in pursuing a Ph.D. in Computer Science and Engineering at the Indian Institute of Technology Guwahati. 

His research focus centers on the exploration of self-supervised techniques for Computer Vision and Deep Learning applications. 
\end{IEEEbiography}
\vspace{-30pt}
\begin{IEEEbiography}
[{\includegraphics[width=1in,height=1.25in,clip,keepaspectratio]{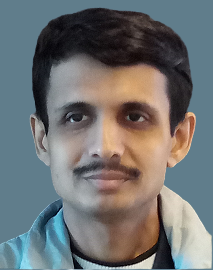}}]{Arijit Sur} received the Ph.D. degree in computer science and engineering from IIT Kharagpur, the M.Sc. degree in computer and information science and the M.Tech. degree in computer science and engineering from the University of Calcutta. He is currently working as a Professor with the Department of Computer Science and Engineering at the Indian Institute of Technology Guwahati. 

His research interest includes Computer Vision using Machine Learning, Steganography and Steganalysis Image and Video Processing, Watermarking, and Adaptive Video Streaming.
\end{IEEEbiography}
\vspace{-30pt}
\begin{IEEEbiography}[{\includegraphics[width=1in,height=1.25in,clip,keepaspectratio]{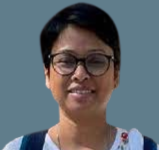}}]{Rashmi Dutta Baruah} received her Ph.D. degree in Communication Systems from Lancaster University, U.K. She is an Assistant Professor in the Department of Computer Science and Engineering, Indian Institute of Technology Guwahati, Assam, India. Currently, she is a CONEX-Plus MARIE CURIE Research Fellow in the Telematics Engineering Department, Universidad Carlos III de Madrid, Spain. 

Her research interest broadly lies in the area of applied Artificial Intelligence and Machine Learning towards providing intelligent solutions to challenging industry-related problems.
\end{IEEEbiography}



\vfill

\end{document}